%% file: main.tex
\pdfoutput=1

\documentclass[11pt]{article}

\usepackage[final]{acl}

\usepackage{times}
\usepackage{latexsym}

\usepackage[T1]{fontenc}

\usepackage[utf8]{inputenc}

\usepackage{microtype}

\usepackage{inconsolata}
\usepackage{amsfonts}
\usepackage{amsmath}
\usepackage{booktabs}
\usepackage{graphicx}
\usepackage{comment}
\usepackage{todonotes}
\usepackage{multirow}
\usepackage{xspace}
\usepackage{xcolor,colortbl}
\usepackage{bm}
\usepackage[normalem]{ulem}
\usepackage{enumitem}
\usepackage{setspace}
\usepackage{url}
\usepackage{supertabular}
\usepackage{array}

\makeatletter
\protected\def\z#1{\pdfsavepos\write\@auxout{\gdef\string\tpos#1{\the\pdflastxpos}}}%
\def\foo#1#2{\ifcsname tpos#1\endcsname\the\dimexpr\csname tpos#2\endcsname sp -\dimexpr\csname tpos#1\endcsname sp\relax\fi}
\makeatother

\newcommand{\eg}{\emph{e.g.,}\ }
\newcommand{\ie}{\emph{i.e.,}\ }
\newcommand{\etc}{\emph{etc.}\ }

\newcommand\blfootnote[1]{%
  \begingroup
  \renewcommand\thefootnote{}\footnote{#1}%
  \addtocounter{footnote}{-1}%
  \endgroup
}

\newcolumntype{x}[1]{>{\centering\arraybackslash\hspace{0pt}}p{#1}}

\title{A Comprehensive Survey of Scientific Large Language Models and \\
Their Applications in Scientific Discovery}

\author{
Yu Zhang\textsuperscript{$\clubsuit$}$^*$,
Xiusi Chen\textsuperscript{$\diamondsuit\clubsuit$}$^*$,
Bowen Jin\textsuperscript{$\clubsuit$}$^*$,
\\
\textbf{Sheng Wang}\textsuperscript{$\heartsuit$},
\textbf{Shuiwang Ji}\textsuperscript{$\spadesuit$},
\textbf{Wei Wang}\textsuperscript{$\diamondsuit$},
\textbf{Jiawei Han}\textsuperscript{$\clubsuit$}
\\ 
\textsuperscript{$\clubsuit$} University of Illinois at Urbana-Champaign \ \ 
\textsuperscript{$\diamondsuit$} University of California, Los Angeles \\
\textsuperscript{$\heartsuit$} University of Washington, Seattle \ \ 
\textsuperscript{$\spadesuit$} Texas A\&M University
\\
{\tt \{yuz9,bowenj4,hanj\}@illinois.edu} \ \ \ 
{\tt \{xchen,weiwang\}@cs.ucla.edu} \\
{\tt swang@cs.washington.edu} \ \ \ 
{\tt sji@tamu.edu}
}

\begin{document}

\maketitle
\begin{spacing}{0.96}
\input{0-Abstract.tex}

\input{1-Introduction.tex}

\input{2-General.tex}

\input{3-Mathematics.tex}

\input{4-Physics.tex}

\input{5-Chemistry.tex}

\input{6-Biology.tex}

\input{7-Geography.tex}

\input{8-Conclusions.tex}
\end{spacing}

\bibliography{emnlp}

\appendix

\setcounter{table}{0}
\renewcommand{\thetable}{A\arabic{table}}

\setcounter{figure}{0}
\renewcommand{\thefigure}{A\arabic{figure}}

\newpage
\onecolumn
\input{9-Appendix.tex}

\end{document}

%% file: 0-Abstract.tex
\begin{abstract}
In many scientific fields, large language models (LLMs) have revolutionized the way text and other modalities of data (\eg molecules and proteins) are handled, achieving superior performance in various applications and augmenting the scientific discovery process.
Nevertheless, previous surveys on scientific LLMs often concentrate on one or two fields or a single modality.
In this paper, we aim to provide a more holistic view of the research landscape by unveiling cross-field and cross-modal connections between scientific LLMs regarding their architectures and pre-training techniques.
To this end, we comprehensively survey over 260 scientific LLMs, discuss their commonalities and differences, as well as summarize pre-training datasets and evaluation tasks for each field and modality.
Moreover, we investigate how LLMs have been deployed to benefit scientific discovery.
Resources related to this survey are available at \url{https://github.com/yuzhimanhua/Awesome-Scientific-Language-Models}.
\blfootnote{$*$ Equal contribution}
\end{abstract}

%% file: 1-Introduction.tex
\section{Introduction}
The emergence of large language models (LLMs) \cite{zhao2023survey} brings a new paradigm to natural language processing (NLP) by replacing specialized models designed for each task with unified models that are reasonably effective for a wide spectrum of problems.
In the scientific domain, such a paradigm not only reshapes people's strategies to handle tasks related to natural language (\eg scientific papers, medical records, and climate reports) but also inspires analogous ideas to deal with other types of data (\eg molecules, proteins, tables, and metadata). 
In addition to understanding existing scientific data, LLMs have shown their potential to accelerate scientific discovery \cite{wang2023scientific,zhang2023artificial,wang2024towards} through generation, planning, \etc

\begin{figure*}[t]
\centering
\includegraphics[width=\linewidth]{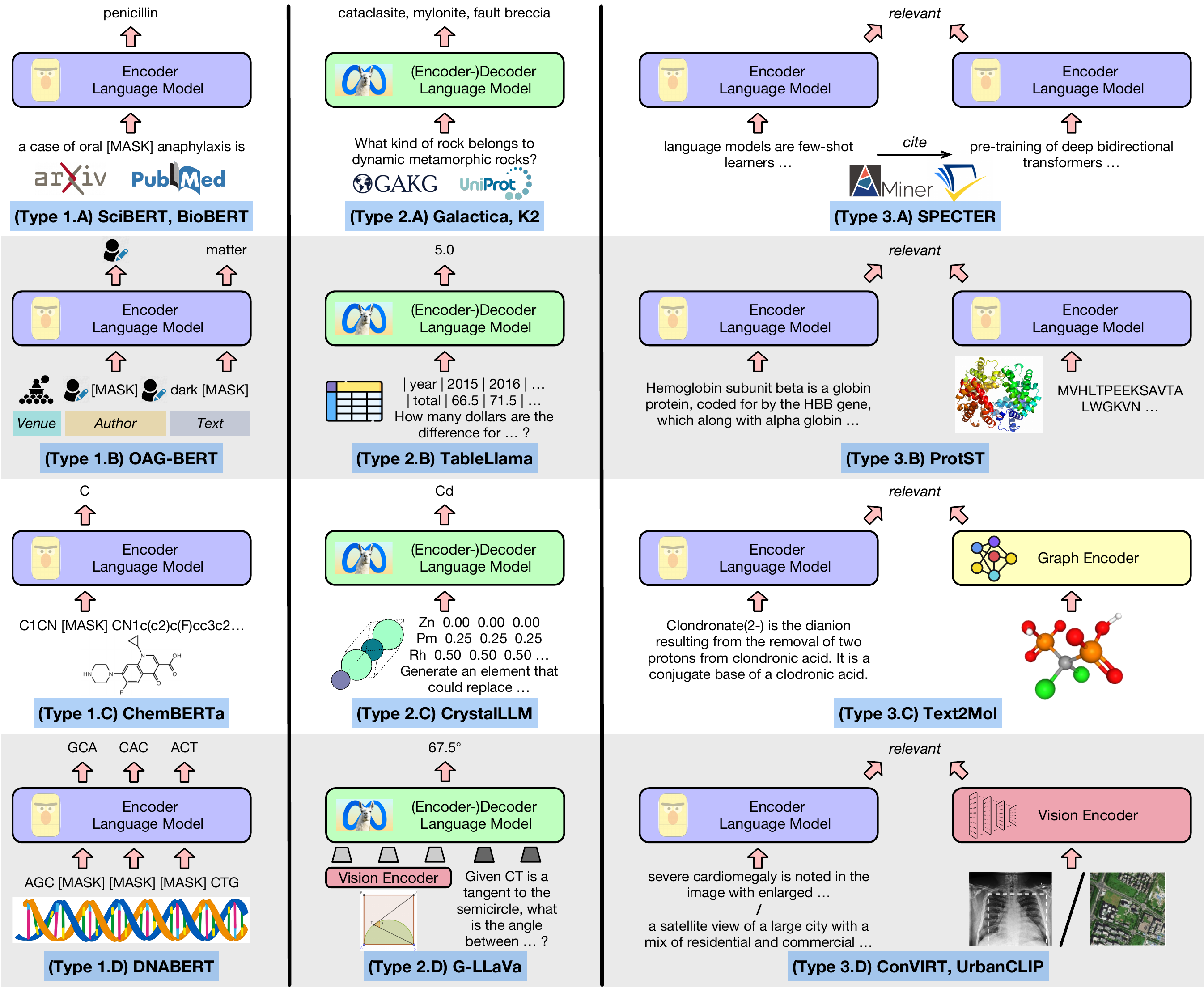}
\caption{Three major types of scientific LLM pre-training techniques. \textbf{(\textsc{Column 1})}: Pre-training encoder LLMs with sequentialized scientific data (\eg text, academic graphs, molecules, biological sequences) via masked language modeling. \textbf{(\textsc{Column 2})}: Pre-training (encoder-)decoder LLMs with sequentialized scientific data (\eg text, tables, crystals, images) via next token prediction (possibly with instruction tuning). \textbf{(\textsc{Column 3})}: Mapping text and relevant sequences/graphs/images closer in the latent space via contrastive learning.}
\label{fig:unification}
\vspace{-0.5em}
\end{figure*}

Given the broad and profound impact of LLMs in various scientific fields across diverse modalities, it becomes necessary to comprehensively review related work in this direction. 
However, existing scientific LLM surveys typically focus on either one or two fields (\eg biomedicine \cite{wang2023pre,he2024foundation,pei2024leveraging,zhang2024scientific} and chemistry \cite{xia2023systematic,pei2024leveraging,zhang2024scientific}) or one modality (\eg text \cite{ho2024survey}) only. 
In fact, if we take a holistic view of the research landscape, we can observe similar and interrelated techniques used to develop LLMs for different fields and modalities.

\autoref{fig:unification} depicts three major types of scientific LLM pre-training strategies (\ie \textsc{Columns 1} to \textsc{3}), for each of which we give 4 examples (\ie \textsc{Types a} to \textsc{d}). 
In \textsc{Column 1}, following BERT \cite{devlin2019bert} and RoBERTa \cite{liu2019roberta}, existing studies use masked language modeling (MLM) to pre-train encoder language models. 
Here, the input can be naturally sequential (\eg papers in each field; protein, DNA, and RNA sequences in the FASTA format \cite{lipman1985rapid}) or artificially linearized (\eg molecules in the SMILES format \cite{weininger1988smiles}; sequences of venue, author, and reference nodes in citation graphs).
In \textsc{Column 2}, inspired by GPT \cite{brown2020language} and LLaMA \cite{touvron2023llama}, previous studies adopt next token prediction to pre-train (encoder-)decoder language models, some of which further adopt instruction tuning and preference optimization \cite{ouyang2022training}. 
Other than plain text input (\eg question-answer pairs from knowledge bases or exams), we see more ways to sequentialize complex scientific data, such as flattening table cells and using particle coordinates to describe crystals. 
Even for images, there are studies in both mathematics \cite{gao2023g} and biomedicine \cite{li2023llava} that exploit a vision encoder to project an image onto several visual tokens and prepend them to text tokens as linearized LLM input.
In \textsc{Column 3}, following DPR \cite{karpukhin2020dense} and CLIP \cite{radford2021learning}, two encoders are pre-trained to map relevant data pairs closer in the latent space via contrastive learning. 
When both modalities are sequential (\eg text-text or text-protein), the model is built upon two LLM encoders. 
When we prefer to keep the non-sequential nature of one modality (\eg molecular graphs \cite{edwards2021text2mol}, chest X-rays \cite{zhang2022contrastive}, and aerial views \cite{yan2024urbanclip}), the corresponding graph or image encoder can be employed.
To summarize, a cross-field, cross-modal survey will more accurately draw the connections between different scientific LLMs, demonstrate their commonalities, and potentially guide their future designs.

\vspace{1mm}
\noindent \textbf{Contributions.} In this paper, motivated by the discussions above, we systematically survey over 260 scientific LLMs encompassing various fields (\eg general science, mathematics, physics, chemistry, materials science, biology, medicine, and geoscience), modalities (\eg language, graph, vision, table, molecule, protein, genome, and climate time series), and sizes (from $\sim$100M to $\sim$100B parameters). 
For each field/modality, we investigate commonly adopted pre-training datasets, model architectures, and evaluation tasks of scientific LLMs. 
Following our motivation, when we discuss model architectures in detail, we link them back to \autoref{fig:unification} to build cross-field cross-modal connections. 
Moreover, we provide a structured summary of these scientific LLMs in \autoref{tab:general}-\autoref{tab:geography} (\autoref{sec:app_table}). 
Furthermore, for different fields, we introduce how LLMs have been deployed to benefit science by augmenting different aspects and stages of the scientific discovery process, such as hypothesis generation, theorem proving, experiment design, drug discovery, and weather forecasting.

%% file: 2-General.tex
\section{LLMs in General Science (\autoref{tab:general})}

\subsection{Language}
The most commonly used pre-training corpora for scientific LLMs are research papers from bibliographic databases, such as AMiner \cite{tang2008arnetminer}, Microsoft Academic Graph (MAG) \cite{sinha2015overview}, and Semantic Scholar \cite{ammar2018construction}. Some of these sources (\eg S2ORC \cite{lo2020s2orc}) contain full-text information of papers, while the others have titles and abstracts only.

The evolution of scientific LLMs bears similarity to that of general-domain LLMs. Specifically, pioneering models utilize paper text in a self-supervised manner during pre-training, aiming to acquire scientific knowledge from large-scale unlabeled corpora. For example, masked language modeling (MLM) is the default pre-training task for scientific LLMs with a BERT backbone (\textsc{Type 1.a} in \autoref{fig:unification}, \eg SciBERT \cite{beltagy2019scibert});
next token prediction is widely used for GPT-based scientific LLMs (\textsc{Type 2.a} in \autoref{fig:unification}, \eg SciGPT \cite{luu2021explaining}). More recently, inspired by the fact that LLMs can be trained to follow natural language instructions \cite{wei2022finetuned,ouyang2022training}, researchers have put more effort into tuning LLMs with instructions to solve complex scientific problems (\textsc{Type 2.a}, \eg Galactica \cite{taylor2022galactica} and SciGLM \cite{zhang2024sciglm}). The instruction tuning data are often derived from datasets for downstream tasks, such as exam question answering \cite{welbl2017crowdsourcing}, and further filtered or augmented by humans or existing LLMs (\eg GPT-4 \cite{achiam2023gpt}).

General scientific LLMs are usually evaluated on common NLP tasks, such as named entity recognition (NER), relation extraction (RE) \cite{luan2018multi}, question answering (QA) \cite{wang2023scibench}, and classification \cite{cohan2019structural}.

\subsection{Language + Graph}
Beyond plain text, scientific papers are associated with rich metadata including venues, authors, and references \cite{zhang2023effect}. Such metadata connect papers into a graph that complements text signals for characterizing paper semantics. To exploit metadata, some studies (\textsc{Type 1.b}, \eg OAG-BERT \cite{liu2022oag}) concatenate paper text with venues/authors as input and perform MLM on both text and metadata; others (\textsc{Type 3.a}, \eg SPECTER \cite{cohan2020specter}) take citation links as supervision and train LLMs to encode linked papers closer in the embedding space. Recent approaches further modify the Transformer architecture in LLMs with Adapters \cite{singh2023scirepeval}, GNN-nested Transformers \cite{jin2023patton}, and Mixture-of-Experts Transformers \cite{zhang2023pre} to better capture graph signals.

Graph-aware scientific LLMs are often evaluated on tasks regarding the relation between two text units (\eg paper-paper or query-paper), including link prediction, retrieval, recommendation, and author name disambiguation. SciDocs \cite{cohan2020specter} and SciRepEval \cite{singh2023scirepeval} are widely adopted benchmark datasets.

\subsection{Applications in Scientific Discovery}
Performant scientific LLMs can work alongside researchers throughout the entire scientific discovery process. Leaving field-specific applications for later sections, here we underscore LLMs' general usefulness in brainstorming and evaluation: \citet{lahav2022search} integrate LLMs into a search engine for the discovery of scientific challenges and directions; \citet{wang2023learning}, \citet{yang2024large}, \citet{baek2024researchagent}, \citet{gu2024generation}, and \citet{si2024can} leverage LLMs to generate novel scientific ideas, directions, and hypotheses on the basis of prior literature and existing knowledge; \citet{zhang2023should} rely on LLMs to find expert reviewers for each submission; \citet{liu2023reviewergpt}, \citet{liang2023can}, and \citet{d2024marg} explore the capacity of GPT-4 to provide useful feedback on research papers to facilitate automatic review generation; \citet{liang2024mapping,liang2024monitoring} also observe the increasing use of LLMs in writing scientific papers and conference peer reviews.

%% file: 3-Mathematics.tex
\section{LLMs in Mathematics (\autoref{tab:mathematics})}

\subsection{Language}
The pre-training text corpora for mathematics LLMs can be categorized into two classes: (1) multiple-choice QA, 
the representative datasets of which include MathQA \cite{amini2019mathqa}, Ape210K \cite{zhao2020ape210k}, and Math23K \cite{wang2017deep}; as well as 
(2) generative QA, 
the representative datasets of which include GSM8K \cite{cobbe2021training}, MATH \cite{hendrycksmath2021}, and MetaMathQA \cite{yu2023metamath}. 

Similarly to general science LLMs, the backbone model of pioneering mathematics LLMs is BERT (\textsc{Type 1.a}, \eg GenBERT \cite{geva2020injecting} and MathBERT \cite{shen2021mathbert}), and these models are mostly trained via MLM.
For GPT-based mathematics LLMs (\textsc{Type 2.a}, \eg GSM8K-GPT \cite{cobbe2021training} and NaturalProver \cite{welleck2022naturalprover}), next token prediction and instruction tuning are major pre-training tasks to generate mathematical proofs and reasoning processes.
The most recent models (\textsc{Type 2.a}, \eg Rho-Math \cite{lin2024rho} and MAmmoTH2 \cite{yue2024mammoth2}) are based on LLaMA and are trained to follow natural language instructions.
However, when an enormous pre-training corpus is available (\eg mathematical web pages and code), next token prediction is still favored as the mere pre-training task \cite{azerbayev2023llemma,lin2024rho} or the companion task \cite{shao2024deepseekmath,ying2024internlm} to build base models.

QA and math world problems (MWP) have been the most common evaluation tasks for mathematics LLMs.
In addition, quantitative reasoning contains more difficult problems, as the model has to provide a complete and self-contained solution without relying on external tools \cite{shao2024deepseekmath,lin2024rho}.
We see a dominance of use from GSM8K and MATH for QA, and from MathQA and Math23K for MWP. For quantitative reasoning, MMLU-STEM \cite{hendrycks2020measuring} and Big-Bench Hard \cite{suzgun2023challenging} are the most widely adopted.

\subsection{Language + Vision}
Geometry is one of the most important branches of mathematics, and it expresses the settings jointly in text and diagrams. As such, it is mandatory to involve the vision modality for geometry LLMs. The most commonly used pre-training datasets for geometry LLMs include Geometry3K \cite{lu2021inter} and GeoQA \cite{chen2021geoqa}, both of which contain multiple-choice geometry problems.

The key to incorporating the vision modality into LLMs is to encode the images and obtain linearized visual representations. Specifically, Inter-GPS \cite{lu2021inter} (\textsc{Type 2.d}) uses RetinaNet \cite{lin2017focal} to transform images into a set of relationships and then applies BART \cite{lewis2020bart} to produce the solution; G-LLaVA \cite{gao2023g} (\textsc{Type 2.d}) encodes visual input via a pre-trained vision Transformer (ViT), concatenates visual embeddings with textual embeddings, and then feeds the concatenation into LLaMA-2 \cite{touvron2023llama2}. These models are by default pre-trained via sequence-to-sequence tasks, where the problem is the input, and the ground-truth answer with optional rationale is the output. Auxiliary loss such as masked image modeling, image construction, or text-image matching, is optionally added for better visual modeling.

Geometry LLMs are evaluated through geometry problem solving, where the model is asked to select the correct answer given the diagram and its caption, the question, and answer options. Renowned evaluation datasets include Geometry3K \cite{lu2021inter}, GEOS \cite{seo2015solving}, and MathVista \cite{lu2023mathvista}.

\subsection{Table}
A large proportion of math knowledge is stored in the form of tabular data. For the ``Table'' modality, notable resources for pre-training include WikiTableQuestions \cite{pasupat2015compositional}, WikiSQL \cite{zhong2017seq2sql}, and WDC Web Table \cite{lehmberg2016large}. 

The challenge in tables is similar to that in diagrams, namely to obtain linearized table representations. 
In most cases, tables are squeezed into linear text sequences as part of the context and are prepended with the question text as the model input.
As one of the first works in this line of research, TAPAS \cite{herzig2020tapas} (\textsc{Type 1.a}) adopts the MLM objective to predict the masked token in both textual and tabular contexts. Recent developments \cite{li2023table,zhang2024tablellm} resemble the design of TableLlama \cite{zhang2023tablellama} (\textsc{Type 2.b}), with LLaMA-2 as the backbone and instruction tuning as the pre-training task. 

Table LLMs are validated through table QA, where the model is asked to produce the correct answer given the table structure, data values, and a question text. Most existing studies have been evaluated on the WikiTableQuestions and WikiSQL datasets. TableInstruct \cite{zhang2023tablellama} is the most recently developed comprehensive benchmark integrating 14 datasets across 11 tasks.

\subsection{Applications in Scientific Discovery}
Mathematics LLMs have great potential to assist humans in
offering potential solutions.
For instance,
AlphaGeometry \cite{trinh2024solving} 
combines an LLM with a symbolic deduction engine, where the LLM generates 
useful constructs
and the symbolic engine applies formal logic to find solutions.
AlphaGeometry solves 25 out of 30 classical geometry problems adapted from the International Mathematical Olympiad.
\citet{sinha2024wu} extend AlphaGeometry by adding Wu's method \cite{chou1988introduction}, further solving 27 out of 30, surpassing human gold medalists.
FunSearch \cite{romera2024mathematical} integrates LLM with program search.
One notable achievement of FunSearch is its ability to find a new solution to the cap set problem in combinatorial optimization.
The solutions generated can be faster and more efficient than those devised by human experts.
In \citet{li2024automated}, LLMs iteratively propose and critique statistical models by leveraging in-context learning and chain-of-thought reasoning \cite{wei2022chain}.

%% file: 4-Physics.tex
\section{LLMs in Physics (\autoref{tab:physics})}
\subsection{Language}
As a derivative of BERT, astroBERT \cite{grezes2021building} (\textsc{Type 1.a}) is further pre-trained using astronomy-related papers via MLM and next sentence prediction. It is evaluated on the NER task.
Likewise, AstroLLaMA \cite{nguyen2023astrollama} (\textsc{Type 2.a}) fine-tunes LLaMA-2 using over 300,000 astronomy abstracts from arXiv. It is evaluated on paper generation and recommendation tasks.
AstroLLaMA-chat \cite{perkowski2024astrollama} (\textsc{Type 2.a}) is the chat version of AstroLLaMA. It is continually trained on a GPT-4 generated domain-specific dialogue dataset.
PhysBERT \cite{hellert2024physbert} (\textsc{Type 1.a}) is the first physics-specific model for sentence embedding trained on a curated corpus of physics literature based on 1.2 million physics papers on arXiv. It is evaluated on physics-tailored tasks, such as information retrieval, classification, and semantic similarity estimation.

\subsection{Applications in Scientific Discovery}
Transformer-based physics LLMs can potentially assist humans in solving differential equations and designing experiments.
For instance, \citet{cai2024transforming} apply Transformer to predict the integer coefficients in the scattering amplitudes of Planar $\mathcal{N}=4$ Super Yang-Mills theory;
RydbergGPT \cite{fitzek2024rydberggpt} uses Transformer to learn the distribution of qubit measurement outcomes that describe an array of interacting Rydberg atoms;
\citet{arlt2024meta} present an initial trial that applies a code-generating LLM to synthesize experimental blueprints for a whole class
of quantum systems in the form of Python code.

%% file: 5-Chemistry.tex
\section{LLMs in Chemistry and Materials Science (\autoref{tab:chemistry})}


\subsection{Language}
LLM pre-training corpora in chemistry and materials science typically come from research papers and databases (\eg Materials Project \cite{jain2013commentary}).
Besides, recent works adopt domain-specific instruction tuning datasets (e.g., Mol-Instructions \cite{fang2023mol} and SMolInstruct \cite{yu2024llasmol}) derived from PubChem \cite{kim2019pubchem}, MoleculeNet \cite{wu2018moleculenet}, \etc

Early studies on chemistry LLMs mostly adopt a moderate-sized encoder-only architecture pre-trained with MLM (\textsc{Type 1.a}, \eg ChemBERT \cite{guo2021automated}, MatSciBERT \cite{gupta2022matscibert}, and BatteryBERT \cite{huang2022batterybert}).
These models are usually evaluated on downstream tasks including reaction role labeling \cite{guo2021automated} and abstract classification \cite{gupta2022matscibert}.
Recently, researchers have focused more on large-scale decoder-only LLMs trained with next token prediction and instruction tuning (\textsc{Type 2.a}).
Examples include ChemDFM \cite{zhao2024chemdfm}, ChemLLM \cite{zhang2024chemllm}, and LlaSMol \cite{yu2024llasmol}.
Given the desired generalization capability of such models, they are evaluated on a diverse set of tasks such as name conversion \cite{kim2019pubchem}, reaction prediction \cite{jin2017predicting}, retrosynthesis \cite{schneider2016s}, text-based molecule design \cite{edwards2022translation}, and crystal generation \cite{antunes2023crystal,flam2023language,gruver2024fine}.

\subsection{Language + Graph}
\label{sec:chem_l+g}
Graphs are appropriate data structures for characterizing molecules \cite{jin2023large}.
Popular datasets containing molecular graphs include ChEBI-20 \cite{edwards2021text2mol,edwards2022translation}, ZINC \cite{sterling2015zinc}, and PCDes \cite{zeng2022deep}.

In some scenarios, molecular graphs appear simultaneously with text information, thus existing works have explored how to encode both effectively.
The first type of such models adopts a GNN as the graph encoder and an LLM as the text encoder.
The two modalities are connected through contrastive learning \cite{liu2023multi} (\textsc{Type 3.c}).
For example, Text2Mol \cite{edwards2021text2mol} uses GCN \cite{kipf2016semi} and SciBERT to encode a molecule and its corresponding natural language description, respectively, for text-to-molecule retrieval.
The second type of such models utilizes an LLM to encode text and graphs simultaneously \cite{zeng2022deep}. Graphs can be either linearized to SMILES strings \cite{edwards2022translation} (\textsc{Type 2.c}) or projected onto virtual tokens with graph encoders \cite{zhao2023gimlet,liu2023molca} (\textsc{Type 2.d}).
For instance, 3D-MoLM \cite{li2024towards} uses a 3-dimensional molecular encoder to represent molecules as tokens and feeds them together with instructions into LLaMA-2 for molecule-to-text retrieval and molecule captioning.

\subsection{Language + Vision}
Complementing text and graph modalities, molecular images form the vision modality in chemistry.
Existing works adopt a similar philosophy to BLIP-2 \cite{li2023blip}, which represents each image as tokens and feeds them into an LLM (\textsc{Type 2.d}).
For example, GIT-Mol \cite{liu2024git} projects all modalities, including graphs and images, into the latent text space and conducts encoding and decoding using T5 \cite{raffel2020exploring}.

\subsection{Molecule}
Different from \autoref{sec:chem_l+g}, this subsection introduces models dealing with molecules without associated text information.
That being said, comparable approaches inspired by LLMs are utilized to develop molecular language models \cite{flam2022language}.
To be specific, most studies adopt SMILES or SELFIES \cite{krenn2020self} strings as the sequential representation of molecules.
Similar to the trend in the ``Language'' modality, pioneering molecular LLMs focus on representation learning with bidirectional Transformer encoders (\textsc{Type 1.c}, \eg SMILES-BERT \cite{wang2019smiles} and MoLFormer \cite{ross2022large}).
For instance, ChemBERTa \cite{chithrananda2020chemberta} adopts the architecture and pre-training strategy similar to those of RoBERTa \cite{liu2019roberta}.
These models exhibit extraordinary abilities in molecular understanding tasks such as molecular property prediction (\eg toxicity classification \cite{wu2018moleculenet} and atomization energy regression \cite{ramakrishnan2014quantum}) as well as virtual screening \cite{riniker2013open}.
Later works explore the idea of representing molecules in an autoregressive fashion (\textsc{Type 2.c}, \eg BARTSmiles \cite{chilingaryan2022bartsmiles} and ChemGPT \cite{frey2023neural}).
For instance, T5Chem \cite{lu2022unified} adopts the T5 backbone and a sequence-to-sequence pre-training objective.
These models are evaluated in generative tasks that include molecule generation \cite{gaulton2017chembl}, reaction prediction, and retrosynthesis.
Besides linearizing molecules, there are studies modifying the Transformer architecture to admit molecular graphs, such as MAT \cite{maziarka2020molecule} and R-MAT \cite{maziarka2024relative}.

\subsection{Applications in Scientific Discovery}
Previous studies have shown that LLMs facilitate autonomous chemical research.
For example, \citet{m2024augmenting} present a chemistry LLM agent, ChemCrow, that can integrate expert-designed tools for organic synthesis, drug discovery, and materials design;
\citet{zheng2023large} demonstrate that LLMs can perform knowledge synthesis from the scientific literature, knowledge inference from data, and interpretable explanation generation in chemistry;
\citet{boiko2023autonomous} develop an LLM-empowered intelligence system, Coscientist, that can design, plan, and perform chemical research.
Moreover, LLMs accomplish complex tasks in chemistry, such as drug and catalyst design and molecular discovery, purely from instructions \cite{white2023future}.
For instance, \citet{ramos2023bayesian} study catalyst and molecule design with in-context learning, removing the requirement for traditional training or simulation processes;
ChatDrug \cite{liu2023chatgpt} explores drug editing using LLMs with a prompt module, a domain feedback module, and a conversation module;
\citet{jablonka2024leveraging} find that fine-tuned LLMs perform comparably to, or even better than, conventional techniques for many chemistry applications, spanning from the properties of molecules and materials to the yield of chemical reactions;
DrugAssist \cite{ye2023drugassist} serves as an LLM-based interactive model for molecule optimization through human-machine dialogue;
\citet{sprueill2023monte,sprueill2024chemreasoner} use LLMs as agents to search for effective catalysts through Monte Carlo Tree Search and the feedback from an atomistic neural network model;
\citet{wang2024efficient} re-engineer crossover and mutation operations for molecular discovery using LLMs trained on extensive chemical datasets.
Meanwhile, benchmarking studies by \citet{mirza2024large} demonstrate that although LLMs achieve superhuman proficiency in many chemical tasks, further research is critical to enhancing their safety and utility in chemical sciences.

%% file: 6-Biology.tex
\section{LLMs in Biology and Medicine (\autoref{tab:biology})}

\subsection{Language}
Besides research articles (\eg titles/abstracts from PubMed \cite{lu2011pubmed} and full text from PMC \cite{beck2003pubmed}),
pre-training corpora for biomedical LLMs include electronic health records (\eg MIMIC-III \cite{johnson2016mimic}, MIMIC-IV \cite{johnson2023mimic}), knowledge bases (e.g., UMLS \cite{bodenreider2004unified}), and health-related social media posts (\eg COVID-19 tweets \cite{muller2023covid}). Recent studies further collect supervised fine-tuning and preference optimization datasets from medical exam questions, knowledge graphs, and doctor-patient dialogues. Examples include ChiMed \cite{ye2023qilin}, MedInstruct-52k \cite{zhang2023alpacare}, and BiMed1.3M \cite{acikgoz2024hippocrates}, many of which have non-English components (\eg Chinese and Arabic).

The watershed moment in the evolution biomedical LLMs is still the emergence of billion-parameter architectures and instruction tuning. Before that, a wide variety of moderate-sized backbones are explored, including both encoder-based (\textsc{Type 1.a}, \eg BioBERT \cite{lee2020biobert}, Bio-ELECTRA \cite{ozyurt2020effectiveness}, BioRoBERTa \cite{lewis2020pretrained}, BioALBERT \cite{naseem2022benchmarking}, and Clinical-Longformer \cite{li2022clinical}) and (encoder-)decoder-based ones (\textsc{Type 2.a}, \eg SciFive \cite{phan2021scifive}, BioBART \cite{yuan2022biobart}, and BioGPT \cite{luo2022biogpt}). Evaluation tasks for these models range from biomedical NER, RE, sentence similarity estimation, document classification, and QA (\ie the BLURB benchmark \cite{gu2021domain}) to natural language inference (NLI) \cite{romanov2018lessons} and entity linking \cite{dougan2014ncbi}. After the watershed, the trend becomes instruction-tuning billion-parameter LLMs (\textsc{Type 2.a}, \eg Med-PaLM \cite{singhal2023large}, MedAlpaca \cite{han2023medalpaca}, and BioMistral \cite{labrak2024biomistral}). Accordingly, evaluation tasks now include single-round QA \cite{jin2021disease,pal2022medmcqa} and multi-round dialogue \cite{wang2023cmb}. Meanwhile, there are studies proposing a Bi-Encoder architecture (\textsc{Type 3.a}, \eg \citet{jin2023medcpt} and \citet{xu2024bmretriever}) that specifically targets biomedical retrieval tasks, the benchmarks of which are NFCorpus \cite{boteva2016full}, TREC-COVID \cite{voorhees2021trec}, \etc 

\subsection{Language + Graph}
Biomedical ontologies capture rich types of relations between entities. Analogously, citation links characterize connections between biomedical papers. Intuitively, jointly leveraging text and such graph information paves the way for multi-hop reasoning in QA. For instance, \citet{yasunaga2022deep} propose to use an LLM and a GNN to encode text and ontology signals, respectively, and deeply fuse them (\textsc{Type 3.c}); \citet{yasunaga2022linkbert} concatenate text segments from two linked papers together and feed the sequence into an LLM for pre-training, which is essentially appending a metadata neighbor (\ie reference) as context for MLM (\textsc{Type 1.b}). Both approaches demonstrate significant improvement in QA tasks that require complex reasoning. 

\subsection{Language + Vision}
\label{sec:bio_l+v}
Biomedical text-image pairs typically come from two sources: (1) medical reports, such as chest X-rays (\eg MIMIC-CXR \cite{johnson2019mimic}) and pathology reports \cite{huang2023visual}; as well as (2) figure-caption pairs extracted from biomedical papers (\eg ROCO \cite{pelka2018radiology} and MedICaT \cite{subramanian2020medicat}).

Most biomedical vision-language models exploit the CLIP architecture \cite{radford2021learning}, where a text encoder and an image encoder are jointly trained to map the paired text and image closer via contrastive learning (\textsc{Type 3.d}). 
The choice of the text encoder evolves from BERT \cite{zhang2022contrastive} and GPT-2 \cite{huang2023visual} to LLaMA \cite{wu2023towards} and LLaMA-2 \cite{liu2023qilin}, while the image encoder evolves from ResNet \cite{huang2021gloria} to ViT \cite{zhang2023biomedclip} and Swin Transformer \cite{thawkar2023xraygpt}. 
MLM, masked image modeling, and text-text/image-image contrastive learning (\ie by creating augmented views within the language/vision modality) are sometimes adopted as auxiliary pre-training tasks.
Besides CLIP, other general-domain vision-language architectures, such as LLaVA \cite{li2023llava}, PaLM-E \cite{tu2024towards}, and Gemini \cite{saab2024capabilities}, have been explored. For instance, LLaVA-Med (\textsc{Type 2.d}) encodes images onto several visual tokens and prepends them to text tokens as the LLM input.
Evaluation tasks of these models encompass image classification, segmentation, object detection, vision QA, text-to-image/image-to-text retrieval, and report generation, the benchmarks of which include CheXpert \cite{irvin2019chexpert}, PadChest \cite{bustos2020padchest}, SLAKE \cite{liu2021slake}, \etc

\subsection{Protein, DNA, RNA, and Multiomics}
The FASTA format \cite{lipman1985rapid} naturally represents proteins as amino acid sequences and DNAs/RNAs as nucleotide sequences, enabling models to treat them as ``languages''.
Representative resources of such sequences include UniRef \cite{suzek2015uniref} and Swiss-Prot \cite{bairoch2000swiss} for proteins, GRCh38 \cite{harrow2012gencode} and the 1000 Genomes Project \cite{consortium2015global} for DNAs, as well as RNAcentral \cite{consortium2019rnacentral} for RNAs.

Encoder-only protein, DNA, and RNA LLMs (\textsc{Type 1.d}), such as ESM-2 \cite{lin2023evolutionary}, DNABERT \cite{ji2021dnabert}, and RNABERT \cite{akiyama2022informative}, adopt BERT-like architectures and MLM as the pre-training task (\textit{i.e.}, predicting masked amino acids, nucleotides, $k$-mers, or codons); decoder-only models, such as ProGen \cite{madani2023large} and DNAGPT \cite{zhang2023dnagpt}, exploit GPT-like architectures and next token prediction as the pre-training task. There are also studies jointly considering text and protein modalities. For instance, ProtST \cite{xu2023protst} matches protein sequences with their text descriptions (\ie names and functions) via contrastive learning (\textsc{Type 3.b}); BioMedGPT \cite{luo2023biomedgpt} first projects proteins onto tokens and then inputs these tokens together with text into LLaMA-2 for instruction tuning, bearing similarity with \textsc{Type 2.d}. 


Existing multiomics LLMs mainly focus on single‐cell transcriptomics (\eg scRNA-seq) data, such as the expression levels of genes within a single cell \cite{franzen2019panglaodb}. Besides BERT-based (\eg Geneformer \cite{theodoris2023transfer}) and GPT-based (\eg scGPT \cite{cui2024scgpt}) architectures, Performer \cite{yang2022scbert,hao2024large} is widely used due to its linear attention complexity in handling long scRNA-seq data.

\subsection{Applications in Scientific Discovery}
Similarly to chemistry, LLMs can automate experiments in biological and medical research. 
For example, CRISPR-GPT \cite{huang2024crispr} augments an LLM agent with domain knowledge to enhance the design process of CRISPR-based gene-editing experiments;
TrialMind \cite{wang2024accelerating} utilizes LLMs to extract results and synthesize clinical evidence from the literature for medical discovery.
Moreover, LLMs can encode biological sequences to capture structural properties and guide protein design.
For instance, ESM-1b \cite{rives2021biological} and ESM-2 \cite{lin2023evolutionary} enable accurate structure prediction of proteins without expensive and time-consuming experiments;
\citet{ferruz2022controllable} fine-tune LLMs on protein families, which can generate highly divergent but still potentially functional novel sequences;
\citet{he2024novo} leverage an LLM for the de novo generation of SARS-CoV-2 antibodies with desired antigen-binding specificity;
\citet{hie2021learning} develop LLMs to evaluate the evolutionary fitness of viral variants using sequence data alone.

%% file: 7-Geography.tex
\section{LLMs in Geography, Geology, and Environmental Science (\autoref{tab:geography})}


\subsection{Language}
Geoscience research papers, climate-related news articles, Wikipedia pages, corporate sustainability reports, knowledge bases (\eg GAKG \cite{deng2021gakg}), and point-of-interest (POI) data (\eg OpenStreetMap \cite{haklay2008openstreetmap}) constitute the pre-training corpora for geoscience LLMs.


Preliminary research on geoscience LLMs focuses on pre-training bidirectional LLMs with the Transformer encoder backbone (\textsc{Type 1.a}, \eg ClimateBERT \cite{webersinke2021climatebert}, SpaBERT \cite{li2022spabert}, and MGeo \cite{ding2023mgeo}).
For instance, SpaBERT and MGeo perform MLM on a sequence of geolocations for geographic entity linking and query-POI matching, respectively.
More recently, related studies concentrate on scaling up decoding-style autoregressive LLMs in geoscience (\textsc{Type 2.a}, \eg K2 \cite{deng2024k2}, OceanGPT \cite{bi2023oceangpt}, and GeoGalactica \cite{lin2024geogalactica}).
For instance, K2 and OceanGPT adapt LLaMA to geoscience and ocean science, respectively, via supervised fine-tuning with domain-specific instructions curated by human experts and/or augmented by general-domain LLMs.
Evaluations of such models are conducted on geoscience benchmarks, such as GeoBench \cite{deng2024k2} and OceanBench \cite{bi2023oceangpt}, which encompass a broad range of tasks including QA, classification, knowledge probing, reasoning, summarization, and generation.

\subsection{Language + Graph}
Some geoscience applications involve graph signals, such as heterogeneous POI networks and knowledge graphs. To handle such signals and text jointly,
ERNIE-GeoL \cite{huang2022ernie} introduces a Transformer-based aggregation layer to deeply fuse text and POI information within a BERT-based architecture;
PK-Chat \cite{deng2023pk} combines an LLM with a pointer generation network on a knowledge graph to build a knowledge-driven dialogue system.

\subsection{Language + Vision}
Aerial views, together with location descriptions, profile urban regions. To address language and vision modalities jointly, UrbanCLIP \cite{yan2024urbanclip} considers the CLIP architecture (\textsc{Type 3.d}), which is also widely adopted by biomedical vision-language models as mentioned in \autoref{sec:bio_l+v}, to perform text-image contrastive learning for urban indicator prediction.

\subsection{Climate Time Series}
The intuitions and methodologies used in LLMs also facilitate the construction of climate foundation models. Based on the ERA5 \cite{hersbach2020era5} and CMIP6 \cite{eyring2016overview} datasets of climate time series, previous studies exploit the ViT and Swin Transformer architectures to pre-train foundation models for weather forecasting. Representative models include FourCastNet \cite{pathak2022fourcastnet}, Pangu-Weather \cite{bi2023accurate}, \etc


\subsection{Applications in Scientific Discovery}
In geography, \citet{wang2023towards} and \citet{zhou2024large} highlight the potential of LLMs in urban planning from sustainability, living, economic, disaster, and environmental perspectives.
In geology, besides climate and weather forecasting, foundation models have been applied to simultaneous earthquake detection and phase picking \cite{mousavi2020earthquake}.
In environmental science, ChatClimate \cite{vaghefi2023chatclimate} enhances GPT-4 by providing access to external, scientifically accurate knowledge on climate change to build a climate science conversational AI.


%% file: 8-Conclusions.tex
\section{Challenges and Future Directions}

In this survey, we compile literature that elucidates the data, architectures, and tasks used for scientific LLM pre-training, as well as how scientific LLMs have been applied to downstream applications in scientific discovery.
In particular, we underscore analogous architectures, tasks, and trends observed during the evolution of scientific LLMs across different fields and modalities.
Beyond reviewing prior research, we present several challenges to inspire further exploration of this topic.

\vspace{1mm}
\noindent\textbf{Diving into Fine-Grained Themes.} Most existing scientific LLMs target a coarse-grained field (\eg chemistry), while some tasks rely on highly specialized knowledge of a fine-grained theme (\eg Suzuki coupling). When LLMs are pre-trained on more general corpora, frequently appeared signals may dominate the model parameter space, and domain-specific tail knowledge may be wiped out. We believe that automatically curating in-depth, theme-focused knowledge graphs \cite{hope2021extracting} to guide the generation process will be a promising direction to tackle this issue.

\vspace{1mm}
\noindent\textbf{Generalizing to Out-of-Distribution Scientific Data.} In the scientific domain, it is common that the testing distribution shifts from the training distribution \cite{zhang2023artificial}: novel scientific concepts keep emerging in newly published papers; unseen molecules with different scaffolds and unseen proteins with different numbers of peptide chains may appear during testing. Handling such out-of-distribution data remains a challenge for pre-trained scientific LLMs. To our knowledge, invariant learning \cite{arjovsky2019invariant} can serve as the theoretical foundation for out-of-distribution analyses, and how to integrate it into LLM pre-training is worth exploring.

\vspace{1mm}
\noindent\textbf{Facilitating Trustworthy Predictions.} LLMs can generate plausible-sounding but factually incorrect output, commonly known as hallucination \cite{ji2023survey}, which is particularly dangerous in high-stakes scientific domains such as chemistry and biomedicine. To mitigate this issue, retrieval-augmented generation (RAG) provides LLMs with relevant, up-to-date, and trustworthy information. However, previous RAG studies in the scientific domain mainly focus on retrieving text \cite{xiong2024benchmarking} and knowledge \cite{jin2024graph}, while scientific data are heterogeneous and multi-modal. We envision that cross-modal RAG (\eg guiding text generation with relevant chemicals and proteins) will present additional opportunities to further enhance the trustworthiness of scientific LLMs.

\section*{Limitations}
This survey primarily covers LLMs in mathematics and natural sciences.
We are aware that LLMs can also significantly impact social sciences by achieving remarkable performance in representative tasks \cite{ziems2024can} and serving as agents for social simulation experiments \cite{horton2023large}, but we leave the survey of these efforts as future work due to space limitations. 
In addition, this paper focuses on LLMs pre-trained on scientific data or augmented with domain-specific knowledge to benefit scientific discovery. There are studies \cite{guo2023can,wang2023scibench,yue2024mmmu,liang2024scemqa} proposing new benchmark datasets of scientific problems but evaluating the performance of general-purpose LLMs only, and we do not include these works in our survey.
Furthermore, some LLMs may belong to more than one field or modality category given our classification criteria in the paper. For instance, BioMedGPT \cite{luo2023biomedgpt} is pre-trained on biology and chemistry data jointly; GIT-Mol \cite{liu2024git} considers the language, graph, and vision modalities simultaneously. For the sake of brevity, we introduce each of them in only one subsection.

\section*{Acknowledgments}
Research was supported in part by US DARPA INCAS Program No. HR0011-21-C0165 and BRIES Program No. HR0011-24-3-0325, National Science Foundation IIS-19-56151, the Molecule Maker Lab Institute: An AI Research Institutes program supported by NSF under Award No. 2019897, and the Institute for Geospatial Understanding through an Integrative Discovery Environment (I-GUIDE) by NSF under Award No. 2118329. Any opinions, findings, and conclusions or recommendations expressed herein are those of the authors and do not necessarily represent the views, either expressed or implied, of DARPA or the U.S. Government.




%% file: 9-Appendix.tex
\section{Summary Tables of Scientific LLMs}
\label{sec:app_table}

\autoref{tab:general}-\autoref{tab:geography} summarize the modality, number of parameters, model architecture, pre-training data, pre-training task(s), and evaluation task(s) of scientific LLMs in each field. Within each field, we categorize models according to their modality; within each modality, we sort models chronologically. To be specific, if a paper has a preprint (\eg arXiv or bioRxiv) version, its publication date is according to the preprint service. Otherwise, its publication date is according to the conference proceeding or journal.

\ 

\ 

\begin{table*}[!h]
\small
\centering
\caption{Summary of LLMs in general science. ``L'': Language; ``L+G'': Language + Graph; ``$\sim$'': generally adopting the architecture but with modifications; ``MLM'': masked language modeling; ``NSP'': next sentence prediction; ``NER'': named entity recognition; ``RE'': relation extraction; ``QA'': question answering.}
\vspace{-0.5em}
\scalebox{0.67}{
\begin{tabular}{lcccccc}
\toprule
\textbf{Model} & \textbf{Modality} & \textbf{Size} & \textbf{Architecture} & \textbf{Pre-training Data} & \textbf{Pre-training Task(s)} & \textbf{Evaluation Task(s)} \\
\midrule

SciBERT \cite{beltagy2019scibert} & L & 110M & BERT & Semantic Scholar & MLM, NSP & NER, RE, classification, parsing \\

\rowcolor{black!10}
SciGPT2 \cite{luu2021explaining} & L & 117M & GPT-2 & S2ORC & next token prediction & paper relationship explanation \\

CATTS \cite{cachola2020tldr} & L & 406M & BART & SciTLDR & sequence to sequence & summarization \\

\rowcolor{black!10}
SciNewsBERT \cite{smeros2021sciclops} & L & 110M & BERT & news headlines & MLM, NSP & scientific claim extraction \\

ScholarBERT \cite{hong2023diminishing} & L & 340M, 770M & BERT & Public.Resource.Org, & MLM & NER, RE, classification \\
& & & & Wikipedia, BookCorpus & & \\

\rowcolor{black!10}
AcademicRoBERTa \cite{yamauchi2022japanese} & L & 125M & RoBERTa & CiNii & MLM & classification, \\
\bottomrule
\end{tabular}
}
\label{tab:general}
\end{table*}

\ 

\begin{table*}[!h]
\small
\centering
\caption{Summary of LLMs in mathematics. ``L+V'': Language + Vision; ``MWP'': math word problems. Other notations have the same meaning as in previous tables.}
\vspace{-0.5em}
\scalebox{0.67}{
\begin{tabular}{p{134.5pt}x{35.1pt}x{47.6pt}x{48.7pt}x{84.6pt}x{116.5pt}x{124.2pt}}
\toprule
\textbf{Model} & \textbf{Modality} & \textbf{Size} & \textbf{Architecture} & \textbf{Pre-training Data} & \textbf{Pre-training Task(s)} & \textbf{Evaluation Task(s)} \\
\midrule

GenBERT \cite{geva2020injecting} & L & 110M & BERT & Wikipedia & MLM, & QA, MWP \\
& & & & & sequence to sequence & \\

\rowcolor{black!10}
MathBERT \cite{shen2021mathbert} & L & 110M & BERT & arXiv, math curricula, & MLM & classification, auto-grading \\
\rowcolor{black!10}
& & & & syllabi, textbooks & & \\

MWP-BERT \cite{liang2022mwp} & L & 110M & BERT & Ape210K & MLM, regression, & QA, MWP \\
& & & & & classification & \\

\rowcolor{black!10}
BERT-TD \cite{li2022seeking} & L & 110M & BERT & Math23K, MathQA & sequence to sequence, & QA, MWP \\
\bottomrule
\end{tabular}
}
\label{tab:mathematics}
\end{table*}

\begin{table*}[!h]
\small
\centering
\scalebox{0.67}{
\begin{tabular}{p{134.5pt}x{35.1pt}x{47.6pt}x{48.7pt}x{84.6pt}x{116.5pt}x{124.2pt}}
\toprule
\multicolumn{2}{l}{\textit{(Mathematics, Table Continued)}} & & & & & \\
\midrule
\textbf{Model} & \textbf{Modality} & \textbf{Size} & \textbf{Architecture} & \textbf{Pre-training Data} & \textbf{Pre-training Task(s)} & \textbf{Evaluation Task(s)} \\
\midrule

Llemma \cite{azerbayev2023llemma} & L & 7B, 34B & LLaMA-2 & Proof-Pile-2 & next token prediction & QA, MWP, quantitative reasoning \\

\rowcolor{black!10}
OVM \cite{yu2023outcome} & L & 7B & LLaMA-2 & GSM8K & supervised fine-tuning & QA, MWP, quantitative reasoning \\
\rowcolor{black!10}
& & 7B & Mistral & & & \\

DeepSeekMath \cite{shao2024deepseekmath} & L & 7B & DeepSeek & math web pages, & next token prediction, & QA, MWP, quantitative reasoning, \\
& & & & instructions & instruction tuning & formal translation \\

\rowcolor{black!10}
InternLM-Math \cite{ying2024internlm} & L & 7B, 20B & InternLM2 & Knowledge Pile, & next token prediction, & QA, MWP, quantitative reasoning, \\
\bottomrule
\end{tabular}
}
\end{table*}

\begin{table*}[!h]
\small
\centering
\caption{Summary of LLMs in physics. Notations have the same meaning as in previous tables.}
\vspace{-0.5em}
\scalebox{0.67}{
\begin{tabular}{p{157.7pt}x{35.1pt}x{33.6pt}x{48.7pt}x{141.8pt}x{77.3pt}x{96.5pt}}
\toprule
\textbf{Model} & \textbf{Modality} & \textbf{Size} & \textbf{Architecture} & \textbf{Pre-training Data} & \textbf{Pre-training Task(s)} & \textbf{Evaluation Task(s)} \\
\midrule

astroBERT \cite{grezes2021building} & L & 110M & BERT & NASA Astrophysics Data System & MLM, NSP & NER \\

\rowcolor{black!10}
AstroLLaMA \cite{nguyen2023astrollama} & L & 7B & LLaMA-2 & arXiv & next token prediction & paper generation, \\
\rowcolor{black!10}
& & & & & & paper similarity estimation \\

AstroLLaMA-Chat \cite{perkowski2024astrollama} & L & 7B & LLaMA-2 & QA pairs, LIMA, OpenOrca, UltraChat & instruction tuning & QA \\

\rowcolor{black!10}
PhysBERT \cite{hellert2024physbert} & L & 110M & BERT & arXiv & MLM, & classification, retrieval, \\
\rowcolor{black!10}
& & & & & contrastive learning & clustering \\

\bottomrule
\end{tabular}
}
\label{tab:physics}
\end{table*}

\begin{table*}[!h]
\small
\centering
\caption{Summary of LLMs in chemistry and materials science. ``L+G+V'': Language + Graph + Vision; ``KG'': knowledge graph; ``SMILES'': simplified molecular-input line-entry system. Other notations have the same meaning as in previous tables.}
\vspace{-0.5em}
\scalebox{0.67}{
\begin{tabular}{p{146.3pt}x{35.1pt}x{47.6pt}x{48.7pt}x{94.5pt}x{90.9pt}x{126.0pt}}
\toprule
\textbf{Model} & \textbf{Modality} & \textbf{Size} & \textbf{Architecture} & \textbf{Pre-training Data} & \textbf{Pre-training Task(s)} & \textbf{Evaluation Task(s)} \\
\midrule

ChemBERT \cite{guo2021automated} & L & 110M & BERT & chemistry journals & MLM & NER \\

\rowcolor{black!10}
MatSciBERT \cite{gupta2022matscibert} & L & 110M & BERT & ScienceDirect & MLM & NER, RE, classification \\

MatBERT \cite{trewartha2022quantifying} & L & 110M & BERT & materials science journals & MLM & NER \\

\rowcolor{black!10}
BatteryBERT \cite{huang2022batterybert} & L & 110M & BERT & Elsevier, Springer, RSC & MLM & QA, classification \\

MaterialsBERT \cite{shetty2023general} & L & 110M & BERT & materials science journals & MLM, NSP & NER \\

\rowcolor{black!10}
Recycle-BERT \cite{kumar2023recycle} & L & 110M & BERT & plastic recycling articles & classification & QA, classification \\

CatBERTa \cite{ock2023catalyst} & L & 125M & RoBERTa & OC20 & regression & regression \\

\rowcolor{black!10}
LLM-Prop \cite{rubungo2023llm} & L & 37M & T5 (encoder) & Materials Project & classification, regression & classification, regression \\

\bottomrule
\end{tabular}
}
\label{tab:chemistry}
\end{table*}

\begin{table*}[!h]
\small
\centering
\scalebox{0.67}{
\begin{tabular}{p{146.3pt}x{35.1pt}x{47.6pt}x{48.7pt}x{94.5pt}x{90.9pt}x{126.0pt}}
\toprule
\multicolumn{2}{l}{\textit{(Chemistry and Materials Science, Table Continued)}} & & & & & \\
\midrule
\textbf{Model} & \textbf{Modality} & \textbf{Size} & \textbf{Architecture} & \textbf{Pre-training Data} & \textbf{Pre-training Task(s)} & \textbf{Evaluation Task(s)} \\
\midrule

ChemDFM \cite{zhao2024chemdfm} & L & 13B & LLaMA & chemistry papers, & next token prediction, & QA, classification, \\
& & & & textbooks, instructions & instruction tuning & name conversion, \\
& & & & & & molecule captioning, \\
& & & & & & text-based molecule design, \\
& & & & & & reaction prediction, retrosynthesis \\

\rowcolor{black!10}
CrystalLLM \cite{gruver2024fine} & L & 7B, 13B, & LLaMA-2 & Materials Project & instruction tuning & crystal generation \\
\bottomrule
\end{tabular}
}
\end{table*}

\begin{table*}[!h]
\small
\centering
\caption{Summary of LLMs in biology and medicine. ``Multi'': Multiomics (\eg single-cell); ``NLI'': natural language inference; ``VQA'': visual question answering; ``EHR'': electronic health record; ``EMR'': electronic medical record; ``PPI'': protein-protein interaction. Other notations have the same meaning as in previous tables.}
\vspace{-0.5em}
\scalebox{0.67}{
\begin{tabular}{p{128.4pt}x{35.1pt}x{48.7pt}x{50.9pt}x{89.3pt}x{114.5pt}x{124.8pt}}
\toprule
\textbf{Model} & \textbf{Modality} & \textbf{Size} & \textbf{Architecture} & \textbf{Pre-training Data} & \textbf{Pre-training Task(s)} & \textbf{Evaluation Task(s)} \\
\midrule

BioBERT \cite{lee2020biobert} & L & 110M, 340M & BERT & PubMed, PMC & MLM, NSP & NER, RE, QA \\

\rowcolor{black!10}
BioELMo \cite{jin2019probing} & L & 93M & ELMo & PubMed & next token prediction, & NER, NLI \\
\rowcolor{black!10}
& & & & & previous token prediction & \\

ClinicalBERT & L & 110M & BERT & MIMIC-III & MLM, NSP & NER, NLI \\
\ \ \ \ \ \cite{alsentzer2019publicly} & & & & & & \\

\rowcolor{black!10}
ClinicalBERT \cite{huang2019clinicalbert} & L & 110M & BERT & MIMIC-III & next token prediction, & word similarity estimation, \\
\rowcolor{black!10}
& & & & & previous token prediction & hospital readmission prediction \\
\bottomrule
\end{tabular}
}
\label{tab:biology}
\end{table*}

\begin{table*}[!h]
\small
\centering
\scalebox{0.67}{
\begin{tabular}{p{128.4pt}x{35.1pt}x{48.7pt}x{50.9pt}x{89.3pt}x{114.5pt}x{124.8pt}}
\toprule
\multicolumn{2}{l}{\textit{(Biology and Medicine, Table Continued)}} & & & & & \\
\midrule
\textbf{Model} & \textbf{Modality} & \textbf{Size} & \textbf{Architecture} & \textbf{Pre-training Data} & \textbf{Pre-training Task(s)} & \textbf{Evaluation Task(s)} \\
\midrule

\rowcolor{black!10}
MMedLM \cite{qiu2024towards} & L & 7B & InternLM & MMedC & next token prediction & QA \\
\rowcolor{black!10}
& & 1.8B, 7B & InternLM2 & & & \\
\rowcolor{black!10}
& & 8B & LLaMA-3 & & & \\

BioMedLM \cite{bolton2024biomedlm} & L & 2.7B & $\sim$GPT-2 & PubMed, PMC & next token prediction & QA \\

\rowcolor{black!10}
Hippocrates \cite{acikgoz2024hippocrates} & L & 7B & LLaMA-2 & PubMed, PMC, & next token prediction, & QA \\
\rowcolor{black!10}
& & 7B & Mistral & medical guidelines, & instruction tuning & \\
\bottomrule
\end{tabular}
}
\end{table*}

\begin{table*}[!h]
\small
\centering
\scalebox{0.67}{
\begin{tabular}{p{128.4pt}x{35.1pt}x{48.7pt}x{50.9pt}x{89.3pt}x{114.5pt}x{124.8pt}}
\toprule
\multicolumn{2}{l}{\textit{(Biology and Medicine, Table Continued)}} & & & & & \\
\midrule
\textbf{Model} & \textbf{Modality} & \textbf{Size} & \textbf{Architecture} & \textbf{Pre-training Data} & \textbf{Pre-training Task(s)} & \textbf{Evaluation Task(s)} \\
\midrule

\rowcolor{black!10}
PLIP \cite{huang2023visual} & L+V & -- & GPT-2 + & Twitter text-image pairs, & text-image matching & classification, \\
\rowcolor{black!10}
& & & ViT & PathLAION & & text-to-image retrieval, \\
\rowcolor{black!10}
& & & & & & image-to-image retrieval \\

MaCo \cite{huang2023enhancing} & L+V & -- & BERT + & MIMIC-CXR & masked image modeling, & classification, segmentation, \\
& & & ViT & & text-image matching & phrase grounding \\

\rowcolor{black!10}
CXR-CLIP \cite{you2023cxr} & L+V & -- & BERT + & MIMIC-CXR, & text-image matching, & classification, \\
\bottomrule
\end{tabular}
}
\end{table*}

\begin{table*}[!h]
\small
\centering
\scalebox{0.67}{
\begin{tabular}{p{128.4pt}x{35.1pt}x{48.7pt}x{50.9pt}x{89.3pt}x{114.5pt}x{124.8pt}}
\toprule
\multicolumn{2}{l}{\textit{(Biology and Medicine, Table Continued)}} & & & & & \\
\midrule
\textbf{Model} & \textbf{Modality} & \textbf{Size} & \textbf{Architecture} & \textbf{Pre-training Data} & \textbf{Pre-training Task(s)} & \textbf{Evaluation Task(s)} \\
\midrule

DNABERT \cite{ji2021dnabert} & DNA & 110M & BERT & GRCh38 & MLM & chromatin profile prediction, \\
& & & & & & promoter prediction, \\
& & & & & & splice site prediction, \\
& & & & & & functional genetic variant \\
& & & & & & identification \\

\rowcolor{black!10}
GenSLMs \cite{zvyagin2023genslms} & DNA & 25M, 250M, & $\sim$GPT-2 & prokaryotic gene & next token prediction & SARS-CoV-2 genome evolution \\
\bottomrule
\end{tabular}
}
\end{table*}

\begin{table*}[!h]
\small
\centering
\caption{Summary of LLMs in geography, geology, and environmental science. ``Climate'': Climate Time Series; ``POI'': point of interest. Other notations have the same meaning as in previous tables.}
\vspace{-0.5em}
\scalebox{0.67}{
\begin{tabular}{p{121.5pt}x{35.1pt}x{48.7pt}x{64.0pt}x{104.5pt}x{110.5pt}x{107.0pt}}
\toprule
\textbf{Model} & \textbf{Modality} & \textbf{Size} & \textbf{Architecture} & \textbf{Pre-training Data} & \textbf{Pre-training Task(s)} & \textbf{Evaluation Task(s)} \\
\midrule

ClimateBERT & L & 82M & DistilRoBERTa & climate-related news, papers, & MLM & classification, fact-checking \\
\ \ \ \ \ \cite{webersinke2021climatebert} & & & & corporate climate reports & & \\

\rowcolor{black!10}
SpaBERT \cite{li2022spabert} & L & 110M, 340M & BERT & OpenStreetMap & MLM, & entity typing, entity linking \\
\rowcolor{black!10}
& & & & & masked entity prediction & \\

MGeo \cite{ding2023mgeo} & L & 213M & $\sim$BERT & text-geolocation pairs & MLM, & query-POI matching \\
& & & & & masked geographic modeling, & \\
& & & & & contrastive learning & \\

\rowcolor{black!10}
K2 \cite{deng2024k2} & L & 7B & LLaMA & geoscience papers, & next token prediction, & QA \\
\rowcolor{black!10}
& & & & Wikipedia, instructions & instruction tuning & \\

OceanGPT \cite{bi2023oceangpt} & L & 7B & LLaMA-2 & ocean science papers, & next token prediction, & QA, classification, extraction, \\
& & & & instructions & instruction tuning & knowledge probing, \\
& & & & & & commonsense reasoning, \\
& & & & & & summarization, generation \\

\bottomrule
\end{tabular}
}
\label{tab:geography}
\end{table*}

\begin{table*}[!h]
\small
\centering
\scalebox{0.67}{
\begin{tabular}{p{121.5pt}x{35.1pt}x{48.7pt}x{64.0pt}x{104.5pt}x{110.5pt}x{107.0pt}}
\toprule
\multicolumn{4}{l}{\textit{(Geography, Geology, and Environmental Science, Table Continued)}} & & & \\
\midrule
\textbf{Model} & \textbf{Modality} & \textbf{Size} & \textbf{Architecture} & \textbf{Pre-training Data} & \textbf{Pre-training Task(s)} & \textbf{Evaluation Task(s)} \\
\midrule

\rowcolor{black!10}
ClimateBERT-NetZero & L & 82M & DistilRoBERTa & Net Zero Tracker & classification & classification \\
\rowcolor{black!10}
\ \ \ \ \ \cite{schimanski2023climatebert} & & & & & & \\

GeoLM \cite{li2023geolm} & L & 110M, 340M & BERT & OpenStreetMap, & MLM, & NER, RE, entity typing, \\
& & & & Wikipedia & contrastive learning & entity linking \\

\rowcolor{black!10}
GeoGalactica \cite{lin2024geogalactica} & L & 30B & Galactica & geoscience papers, code, & next token prediction, & QA, knowledge probing, \\
\rowcolor{black!10}
& & & & Wikipedia, instructions & instruction tuning & quantitative reasoning, \\
\bottomrule
\end{tabular}
}
\end{table*}